\documentclass[11pt]{article}

\usepackage[]{acl}
\usepackage{times}
\usepackage{latexsym}

\usepackage[T1]{fontenc}
\usepackage[utf8]{inputenc}

\usepackage{microtype}
\usepackage{xspace,mfirstuc,tabulary}
\usepackage{booktabs}
\usepackage{amssymb}
\usepackage{pifont}
\usepackage{amsmath}
\usepackage{multirow,booktabs, hhline}
\usepackage[ruled,noend]{algorithm2e}
\usepackage{arydshln}
\usepackage{amsmath, bm}
\newcommand\ourmodel{WebCPM\xspace}
\usepackage{graphicx}
\usepackage{color}
\usepackage{bbm}
\usepackage{bbding}
\usepackage{subfigure}
\usepackage{makecell}
\usepackage{CJKutf8}
\usepackage{cleveref}
\usepackage{listings}
\crefname{section}{§}{§§}
\Crefname{section}{§}{§§}

\definecolor{lightergray}{RGB}{230,230,230}
\definecolor{DarkGreen}{RGB}{30,130,30}
\newcommand{\cmark}{\textcolor{DarkGreen}{\ding{51}}}
\newcommand{\xmark}{\textcolor{red}{\ding{55}}}%

\usepackage{enumitem}

\newenvironment{itemize*}%
 {\leftmargini=20pt\begin{itemize}%
  \setlength{\itemsep}{3pt}%
  \setlength{\parskip}{0pt}%
  }%
 {\end{itemize}}
\newenvironment{enumerate*}%
 {\begin{enumerate}%
  \setlength{\itemsep}{0pt}%
  \setlength{\parskip}{0pt}}%
 {\end{enumerate}}

\title{\textsc{\ourmodel}: \\ Interactive Web Search for Chinese Long-form Question Answering}

\author{
 Yujia~Qin$^{1}$, Zihan~Cai$^{1}$, Dian~Jin$^{1}$, Lan~Yan$^{1}$, Shihao~Liang$^{3}$, Kunlun~Zhu$^{3}$, \\  \textbf{Yankai~Lin}$^{2*}$, \textbf{Xu~Han}$^{1}$, \textbf{Ning~Ding}$^1$, \textbf{Huadong~Wang}$^1$, \textbf{Ruobing Xie}$^4$, \textbf{Fanchao Qi}$^1$, \\ \textbf{Zhiyuan Liu}$^{1}\thanks{\ \  Corresponding author.}$\hspace{0.4em}, \textbf{Maosong Sun}$^{1*}$, \textbf{Jie Zhou}$^{4}$ \\
 $^1$NLP Group, DCST, IAI, BNRIST, Tsinghua University, Beijing \\
 $^2$Gaoling School of Artificial Intelligence, Renmin University of China, Beijing \\
 $^3$ModelBest Inc.
 $^4$Pattern Recognition Center, WeChat AI, Tencent Inc. \\
\texttt{qyj20@mails.tsinghua.edu.cn}\\
}

\begin{document}
\maketitle

\begin{abstract}
Long-form question answering (LFQA) aims at answering complex, open-ended questions with detailed, paragraph-length responses. The de facto paradigm of LFQA necessitates two procedures: information retrieval, which searches for relevant supporting facts, and information synthesis, which integrates these facts into a coherent answer. In this paper, we introduce \ourmodel, the first Chinese LFQA dataset. One unique feature of \ourmodel is that its information retrieval is based on interactive web search, which engages with a search engine in real time. Following WebGPT~\citep{nakano2021webgpt}, we develop a web search interface. We recruit annotators to search for relevant information using our interface and then answer questions. Meanwhile, the web search behaviors of our annotators would be recorded. In total, we collect $5,500$ high-quality question-answer pairs, together with $15,372$ supporting facts and $125,954$ web search actions. We fine-tune pre-trained language models to imitate human behaviors for web search and to generate answers based on the collected facts. Our LFQA pipeline, built on these fine-tuned models, generates answers that are no worse than human-written ones in $32.5\%$ and $47.5\%$ of the cases on our dataset and DuReader~\citep{he-etal-2018-dureader}, respectively. The interface, dataset, and codes are publicly available at \url{https://github.com/thunlp/WebCPM}.
\end{abstract}
\section{Introduction}
Long-form question answering (LFQA)~\citep{fan-etal-2019-eli5} targets answering complex, open-ended questions with detailed, paragraph-length responses. Current LFQA solutions generally follow the \textit{retrieve-then-synthesize} paradigm, which comprises two core ingredients: information retrieval and information synthesis. The former searches external knowledge sources (e.g., the web) for diverse relevant supporting facts, and the latter integrates the collected facts into a coherent answer.

\begin{table*}[!t]
    \centering
    \small
    \resizebox{\linewidth}{!}{
    \begin{tabular}{lcccccc}
        \toprule
        \textbf{Resource} & \makecell{\textbf{\ourmodel}\\(this work)}& \makecell{ \textbf{DuReader} \\\citep{he-etal-2018-dureader} } & 
        \makecell{ \textbf{CMRC}\\\citep{cui-etal-2019-span} } &  \makecell{ $\textbf{C}^3$\\\citep{sun-etal-2020-investigating} } & \makecell{ \textbf{WebGPT}\\\citep{nakano2021webgpt} } & \makecell{\textbf{GopherCite}\\\citep{menick2022teaching}}
        \\
         \cmidrule(lr){1-1}  \cmidrule(lr){2-2}  \cmidrule(lr){3-3}  \cmidrule(lr){4-4}  \cmidrule(lr){5-5} \cmidrule(lr){6-6}  \cmidrule(lr){7-7}
          Language? & ZH & ZH & ZH & ZH & EN & EN \\ 
          Is Public?   & \cmark & \cmark & \cmark & \cmark & \xmark & \xmark \\ 
          Targets long-form QA? & \cmark & \xmark & \xmark & \xmark & \cmark & \cmark \\ 
          Has free-form answer? & \cmark & \cmark & \xmark & \xmark & \cmark & \cmark \\ 
          Has web search behavior?   & \cmark & \xmark & \xmark & \xmark & \cmark & \xmark \\ 
          \hdashline
          Avg. question length   & $\mathbf{29.0}$ & $9.6$ & $16.3$ & $12.2$ & -- & -- \\ 
          Avg. supporting fact length  & $\mathbf{555.7}$ & $187.3$ & $495.5$ & $116.9$ & -- & -- \\ 
          Avg. answer length   & $\mathbf{257.5}$ & $104.9$ & $17.0$ & $5.5$ & -- & -- \\ 
         \bottomrule
    \end{tabular}
    }
    \caption{A comparison of our \ourmodel to relevant datasets. ``--'' means the information is unknown. For the length statistics, we record the number of Chinese characters.
    }
    \label{tab:dataset_comparison}
\end{table*}

One defect of the conventional LFQA paradigm is that it often resorts to \textit{non-interactive} retrieval methods, which use the original question as the query to retrieve a pile of uncurated information. On the contrary, humans are able to perform \textit{interactive web search} by engaging with a search engine in real time. For a complex question, humans tend to decompose it into multiple sub-questions and ask them in sequence. By identifying and browsing relevant information, humans can improve their understanding of the topic and refine their searches by asking follow-up questions or related terms. This iterative process enables expanding the scope of their searches and improving the results they receive. Overall, interactive web search not only provides access to diverse information sources, but also reflects the cognitive process of how humans solve questions, which allows for better interpretability.

WebGPT~\citep{nakano2021webgpt} is one pioneering work that supports interactive web search for LFQA. The authors first build a web search interface backed up by Microsoft Bing, then recruit annotators to collect information using the interface to answer questions. After that, they fine-tune GPT-3~\citep{NEURIPS2020_1457c0d6} to imitate human behaviors for web search and to organize the collected information into answers. In the experiments, WebGPT shows exceptional ability in LFQA, even surpassing human experts. Despite its impressive performance, WebGPT still remains mysterious to the community. This is because WebGPT's interface, dataset, and trained models are not publicly available, and the inner workings of its core design elements remain opaque. These factors make it hard for the community to understand the challenges of interactive web search for LFQA and to continue exploring this line of study. 

\begin{figure*}[!t]
    \centering
    \subfigure{\includegraphics[width=0.95\textwidth]{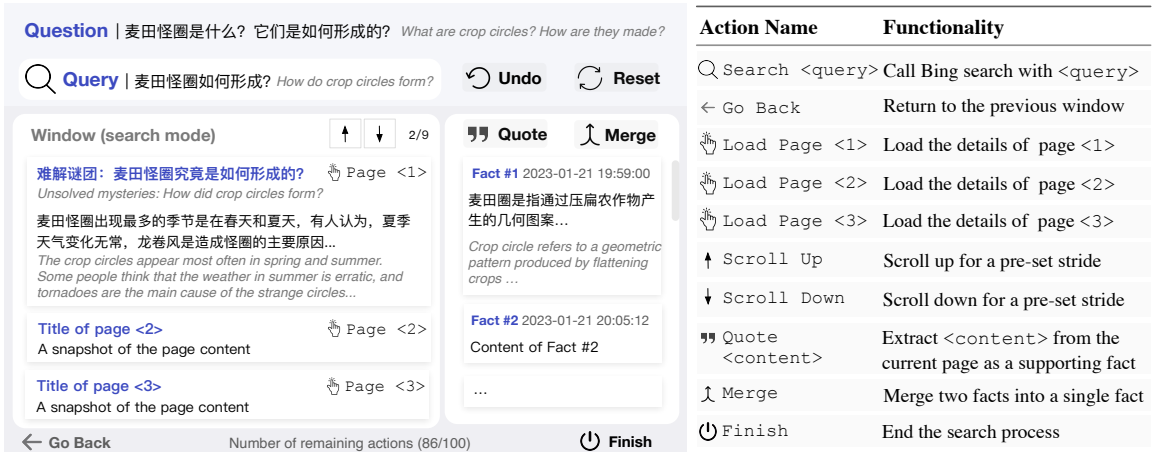}}
    \caption{Left: an example screenshot of our interface in the \textit{search mode}. Right: the actions our interface supports.}
    \label{fig:interface}
\end{figure*}

In view of this, we deem it urgent to provide an accessible platform and public benchmark for this area. To this end, we first construct an interface (Figure~\ref{fig:interface}) to record web search behaviors when humans gather relevant information for long-form questions. In the interface, users can execute pre-defined actions to perform multiple rounds of searching and browsing. When finding relevant information on a web page, they can record it as a supporting fact. Meanwhile, their web-browsing behaviors will be recorded. After collecting enough information, users can finish the web search and answer the questions based on their collected facts.

Based on the interface, we choose Chinese as the testbed and construct \textbf{\ourmodel}, focusing on interactive \textbf{Web} search with \textbf{C}hinese \textbf{P}re-trained \textbf{M}odels. \ourmodel is the first public QA dataset that involves interactive web search, and also the first dataset that targets Chinese LFQA. \ourmodel contains $5,500$ question-answer pairs, together with $15,372$ supporting facts and $125,954$ web search actions. Table~\ref{tab:dataset_comparison} summarizes the difference between \ourmodel and relevant QA datasets. Among existing Chinese QA datasets, \ourmodel possesses the longest question, supporting fact, and answer, which shows the complexity of the questions and the richness of the annotated answers.

Then we propose a general framework consisting of (1) a \textit{search model}, which imitates human web search behaviors for information retrieval. Specifically, the search model comprises three modules to execute a series of pre-defined actions on our interface: an action prediction module, a search query generation module, and a supporting fact extraction module; (2) a \textit{synthesis model}, which generates a coherent answer conditioned on the collected facts.

In the experiments, we choose $8$ representative pre-trained language models (PLMs) with up to $10$B parameter size, and evaluate their ability of interactive web search and information synthesis. We find that scaling model sizes is critical to achieving better performance. By selecting the best-performing backbone PLM for the search and synthesis model, we combine them into a holistic LFQA pipeline and compare its capability with humans. Human evaluation reveals that our pipeline generates answers that are no worse than humans $32.5\%$ of the time on our test set. When applied to questions whose annotated answers are longer than $400$ Chinese characters from DuReader~\citep{he-etal-2018-dureader}, our pipeline generates answers that are better than golden annotated ones $47.5\%$ of the cases, showing satisfying out-of-distribution generalization performance. We also show that our search model surpasses the conventional non-interactive retrieval method. Finally, we analyze the contribution of core design elements of our framework and the human-like behaviors our models acquire. We envision these resources to serve as the testbed for other research topics, such as behavior cloning~\citep{bain1995framework} and tool learning~\citep{qin2023tool}.
\section{Related Work}

\paragraph{Retrieval and Synthesis in LFQA.} 
For information retrieval, prior works generally resort to local repositories (e.g., Wikipedia). Recently there is a surge of interest in leveraging the whole web as the knowledge source~\citep{nakano2021webgpt,lazaridou2022internet,menick2022teaching,thoppilan2022lamda}, which not only widens the scope of information sources but enables real-time coverage of up-to-date knowledge. 
On the other hand, how to structure the retrieved facts into a plausible and nuanced answer for LFQA is still under-explored. Some investigated how humans craft complicated answers, either by studying the functional structures of long-form answers~\citep{xu-etal-2022-answer} or exploring how to compose exemplification in answers~\citep{wang-etal-2022-modeling}; others revisit existing evaluation metrics of LFQA~\citep{krishna-etal-2021-hurdles}.

\paragraph{Comparison with WebGPT.}
We largely follow WebGPT and also propose improved design elements (with details elaborated in \cref{sec:difference}), including (1) \textit{interface}: we modify the actions defined by WebGPT to make them easier for model learning and more user-friendly; (2) \textit{framework}: we decompose web search into $3$ sub-tasks and implement a modular search model. We additionally explore how to teach the synthesis model to ignore irrelevant facts (\cref{sec:in_depth_analysis}) and generate novel contents (\cref{sec:novel_generation}); (3) \textit{evaluation and analysis}: besides evaluating \textbf{the whole} pipeline following WebGPT (\cref{sec:exp_pipeline}), we also evaluate each \textbf{individual} module (\cref{sec:exp_sub_task} and \cref{sec:in_depth_analysis}). This fine-grained evaluation helps us better understand the contribution of core design elements of our framework and the human behaviors learned by our model.

\paragraph{Tool Learning.}
Recent research demonstrates PLMs with promising capabilities of manipulating tools, i.e., tool learning~\citep{qin2023tool}. PLMs can make sequential decisions in complex interactive environments, such as planning in robotic tasks~\citep{huang2022language,saycan2022arxiv,huang2022inner}, manipulating search engines~\citep{nakano2021webgpt}, shopping on e-commerce websites~\citep{yao2022webshop}, etc. By harnessing the rich world knowledge learned during pre-training, PLMs can perform grounded actions to interact with the real world. We envision our benchmark to serve as the testbed for future explorations in this area.
\section{Web Search Environment}
Following WebGPT, we construct a text-only interface to record web search behaviors when humans gather relevant information for long-form questions. Our interface, backed up by \href{https://www.microsoft.com/en-us/bing/apis/bing-web-search-api}{Bing Search API}, supports $10$ mainstream web search actions as shown in Figure~\ref{fig:interface}. When an action is executed, our interface responds with changes in the window.

When the action \texttt{Search} is performed, the interface enters \textit{search mode} (Figure~\ref{fig:interface}), which displays the links recommended by Bing for a specific query \texttt{<query>}. Each link comprises a title and a brief snapshot of the specific web page. Each window displays three links one time, and more links can be accessed by executing the \texttt{Scroll Down} action.

When finding the $i$-th link in the current window to be relevant, users could execute the \texttt{Load Page <i>} action ($i \!\in\! \{1,2,3\}$). The interface would enter the \textit{browsing mode} (Figure~\ref{fig:browsing_mode} in the appendix) and render the texts cleaned from the HTML of the $\texttt{<i>}$-th web page. The content users could view at a time in the window is restricted up to $500$ Chinese characters, and more content can be accessed with the \texttt{Scroll} action. Users can utilize the \texttt{Quote} action to extract consecutive sentences in the current window as a supporting fact. To enable extracting texts that stretch across two windows, the \texttt{Merge} action is designed to merge the last two facts into a single fact (see \cref{sec:website_front_end} for more details). We also display all the existing extracted supporting facts for users.

After browsing the $i$-th page, users can return to the previous \textit{search mode} using the \texttt{Go Back} action to access other links. Meanwhile, a refined query can be sent at any time. In general, users can freely interact with our interface multiple times until executing the \texttt{Finish} action or triggering the maximum number of actions ($100$ in our case). The interface would automatically record meaningful actions and observations during web search. Owing to the multilingual nature of Bing system, although this work focuses on Chinese, our interface can be flexibly adapted to other languages as well. For more technical details, please refer to \cref{sec:interface_detail}.
\section{Data Collection}
\label{sec:qa_data}

We employ $23$ \textbf{annotators} from different walks of life, who are experienced in search engine operation. We ask them to answer long-form questions by first searching for relevant information using our interface, then writing a nuanced answer. For quality control, we recruit $8$ experts familiar with QA research as \textbf{quality inspectors}. Next, we introduce the construction process of our dataset, with detailed annotation guides left in \cref{sec:annotation_principle}.

\paragraph{Question Creation.}
Creating new long-form questions from scratch without any reference is counterproductive, thus we turn to public QA forums as the question source. Specifically, we engage annotators to refer to the questions on an English QA forum \href{https://www.reddit.com/r/explainlikeimfive}{Reddit}, and then create new questions written in Chinese. The details of this creation process are elaborated in \cref{sec:data_collection_detail}. We find empirically that questions created in this way often necessitate multiple rounds of searching and browsing to collect sufficient information.

\paragraph{Interactive Web Search.}
Given a question, we ask annotators to search for accurate and relevant information from trusted sources using our interface. This process may involve sending refined queries to Bing multiple times, as well as exploring various web pages they deem to be relevant. We require annotators to carefully judge the factual accuracy of the information before extracting it as a supporting fact. The search process would be finished until sufficient supporting facts are collected. Among our created questions, $26.2$\% are unanswerable and finally discarded because annotators cannot find sufficient useful information.

\paragraph{Answer Annotation.}
After gathering enough supporting facts, the annotators would write self-contained answers based on their collected information. We give them instructions for answer annotation, including writing answers that are relevant to the question and have rich content, maintaining logical consistency, clarity, and coherence, and providing viewpoints in an unbiased manner.

\begin{figure*}[t!]
  \footnotesize
  \rule{\linewidth}{1pt}
    \noindent \textbf{Question:} \\
    \begin{CJK}{UTF8}{gbsn}
  麦田怪圈是什么？它们是如何形成的？
\end{CJK} \\[1.7mm]
\noindent \textbf{Translated Question:} \\
    \noindent 
    What are crop circles? How are they made?
    \\[1.7mm]
\noindent \textbf{Human Action Sequence:} \\
  $\texttt{Search} \!\rightarrow\! \texttt{Load Page <1>} \!\rightarrow\! \texttt{Quote} \!\rightarrow\! \texttt{Scroll Down} \times 5 \!\rightarrow\! \texttt{Scroll Up} \!\rightarrow\! \texttt{Scroll Down} \times 11 \!\rightarrow\! \texttt{Go Back} \!\rightarrow\! \texttt{Search} \!\rightarrow\! \texttt{Load Page <1>} \!\rightarrow\! \texttt{Go Back} \!\rightarrow\! \texttt{Load Page <3>} \!\rightarrow\! \texttt{Scroll Down} \times 4 \!\rightarrow\! \texttt{Scroll Up} \times 3 \!\rightarrow\! \texttt{Quote} \!\rightarrow\! \texttt{Scroll Down} \!\rightarrow\! \texttt{Quote} \!\rightarrow\! \texttt{Merge} \!\rightarrow\! \texttt{Quote} \!\rightarrow\! \texttt{Scroll Down} \!\rightarrow\! \texttt{Quote} \!\rightarrow\! \texttt{Finish}$
  \\[1.7mm]
    \noindent \textbf{Supporting Facts:} \\ 
    \begin{CJK}{UTF8}{gbsn}
1. 麦田怪圈（Crop Circle），是指在麦田或其它田地上，通过某种未知力量（大多数怪圈是人类所为）把农作物压平而产生出来的几何图案。这个神秘现象有时被人们称之为“Crop Formation”。麦田怪圈的出现给了对支持外星人存在论的人们多种看法。
   
2. 人为说：人为说一般认为，麦田圈是用木板压成的。木板两头系上绳子形成圈套，在制作时，一脚踩在木板上拖动木板压倒麦子，并拉着细绳与圆心保持固定的距离，逐渐就可以形成一个圆圈。为了便于制造，主要形状所有圆圈的直径都可以被6除尽。以前曾经出现过制作麦田圈被当场抓获的事情，制作者使用的就是这种工具。

3. 自然形成说：也有人认为，麦田圈只是一种，成因还未被人类发现。就像雷电，古时候人类也是以为是雷神电母做的，对于麦田圈中经常出现人文信息的现象，他们认为这只是人们“先入为主”造成的错觉。

4. 磁场说：有专家认为，磁场中有一种神奇的移动力，可产生一股电流，使农作物“平躺”在地面上。美国专家杰弗里·威尔逊研究了130多个麦田怪圈，发现90\%的怪圈附近都有连接高压电线的变压器，方圆270米内都有一个水池。由于接受灌溉，麦田底部的土壤释放出的离子会产生负电，与高压电线相连的变压器则产生正电，负电和正电碰撞后会产生电磁能，从而击倒小麦形成怪圈。
\end{CJK}
     \\[1.7mm]
    \noindent \textbf{Translated Supporting Facts:} \\
    \noindent
1. Crop circles, also known as “crop formations”, are geometric patterns that appear in fields of crops, often as a result of unknown forces (though most crop circles are actually created by humans). This mysterious phenomenon has sometimes been used to support the belief in extraterrestrial life.

2. Human-made theory: According to the human-made theory, crop circles are believed to be created using planks of wood. The ends of the planks are attached to ropes to form a loop, and the crops are flattened by dragging the planks across them while maintaining a fixed distance from the center of the circle using the ropes. This process can gradually create a circular pattern. To make the process easier, the diameters of most circles can be evenly divided by 6. In the past, there have been instances of crop circle makers being caught in the act, and they have used this method to create the circles.

3. Natural formation theory: According to the natural formation theory, some people believe that crop circles are a phenomenon whose cause has not yet been discovered by humans. Similar to how lightning was once thought to be the work of the god, the appearance of human-like symbols in crop circles is seen as a result of people's preconceptions.

4. Magnetic field theory: According to the magnetic field theory, some experts believe that crop circles are created by a mysterious force in the magnetic field that generates an electric current that causes the crops to lie flat on the ground. American expert Jeffrey Wilson studied over 130 crop circles and found that 90\% were located near transformers connected to high-voltage power lines and within 270 meters of a pond or other source of water. The irrigation of the soil in the fields can cause the release of ions in the soil at the bottom of the field, creating a negative charge. The transformer connected to the high-voltage power lines generates a positive charge, and the collision of the negative and positive charges creates electromagnetic energy that knocks down the wheat, forming a crop circle.
 \\[1.7mm]
    \noindent \textbf{Answer:} \\
    \begin{CJK}{UTF8}{gbsn}
麦田怪圈是指在麦田或其它田地上，通过某种未知力量（大多数怪圈是人类所为）把农作物压平而产生出来的几何图案。【1】而麦田怪圈的形成有三个说法，人为说、自然说与磁场说。人为说一般认为，麦田圈是用木板压成的。木板两头系上绳子形成圈套，在制作时，一脚踩在木板上拖动木板压倒麦子，并拉着细绳与圆心保持固定的距离，逐渐就可以形成一个圆圈。为了便于制造，主要形状所有圆圈的直径都可以被6除尽。【2】自然形成说认为麦田怪圈的成因还未被人类发现。就像雷电，古时候人类也是以为是雷神电母做的，对于麦田圈中经常出现人文信息的现象，他们认为这只是人们“先入为主”造成的错觉。【3】磁场说认为，磁场中有一种神奇的移动力，可产生一股电流，使农作物“平躺”在地面上。【4】
\end{CJK} \\[1.7mm]
    \noindent \textbf{Translated Answer:} \\
Crop circles are geometric patterns created by flattening crops through an unknown force (most crop circles are made by humans) in a field of wheat or other crops. [1] There are three theories about the formation of crop circles: human-made, natural, and magnetic fields. The human-made theory generally believes that crop circles are made using wooden boards. The wooden boards are connected by ropes to form a hoop, and during the production process, one-foot steps on the wooden board, dragging the wooden board to flatten the wheat, and pulling the fine rope to maintain a fixed distance from the center, gradually forming a circle. In order to facilitate manufacture, the main shape of all circles has a diameter that can be evenly divided by 6. [2] The natural formation theory believes that the cause of crop circles has not yet been discovered by humans. Like lightning, ancient humans also thought it was made by the god, and for the phenomenon of human information often appearing in crop circles, they think it is just a ``preconceived'' illusion caused by people. [3] The magnetic field theory believes that there is a mysterious moving force in the magnetic field that can generate an electric current, causing crops to ``lie flat'' on the ground. [4]
    \\[-1.5mm]
  \rule{\linewidth}{1pt}
  \caption{A sampled example from \ourmodel, where we translated the original Chinese version into English.}
  \label{fig:example_answers_1}
\end{figure*}

\paragraph{Quality Control.}
Each annotated instance is checked and approved by the quality inspectors before being selected for the final dataset. First, inspectors would manually inspect the action sequences recorded on the interface and discard low-quality ones (e.g., those with evident clerical errors in the issued queries). Second, they would carefully check the collected supporting facts. If these facts are apparently insufficient to answer the question, irrelevant to the question, or factually incorrect, the corresponding action sequence would be abandoned. 
The above procedures remove $25\%$ collected instances. For the remaining instances, inspectors would carefully examine their annotated answers. If an answer contradicts the abovementioned instructions, inspectors would return it to annotators and point out which requirement is not satisfied. Annotators would revise their answers possibly for multiple rounds until the revised answer is up to standard.

\begin{figure*}[!t]
    \centering
    \subfigure{\includegraphics[width=0.95\textwidth]{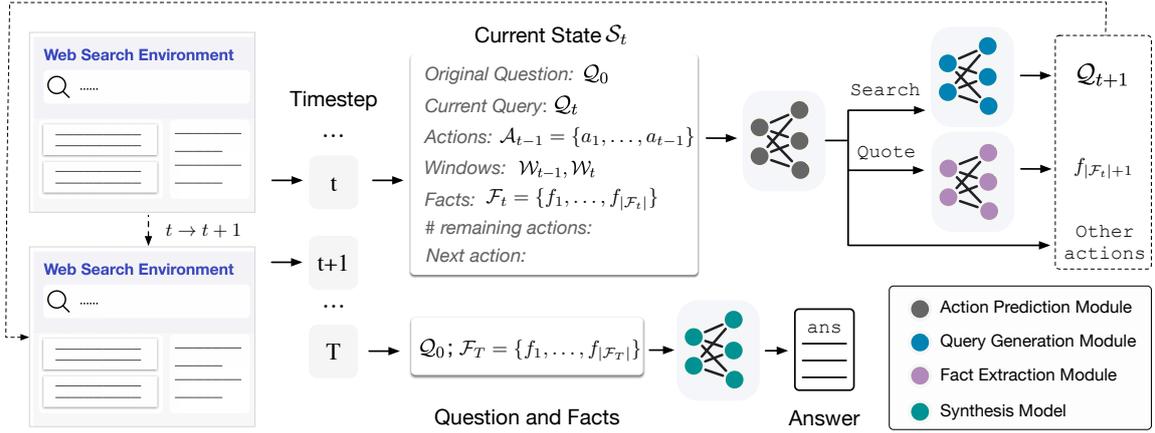}}
    \caption{Illustration of our LFQA framework. For an action sequence of T steps, the search model (consisting of $3$ modules) executes actions to collect supporting facts, which are sent to the synthesis model for answer generation.}
    \label{fig:method}
\end{figure*}

\paragraph{Dataset Statistics.}
Ultimately, we collect $5,500$ instances, each formatted in a tuple of (\textit{question}, \textit{web search behavior}, \textit{supporting fact}, \textit{answer}), and also record the observations at each action execution. We display an example in Figure~\ref{fig:example_answers_1} for reference, where we present the following: the original question, the simplified action sequence, the collected supporting facts, and the annotated answer. We partition the dataset into \{$4,700$, $400$, $400$\} as the training, development, and test set. On average, each question involves performing $22.9$ actions, sending $2.5$ queries, and loading $3.3$ web pages. The detailed proportion of each action is visualized in Figure~\ref{fig:action_pie} in the appendix.

\section{Framework}
In this section, we introduce how to teach PLMs for (1) interactive web search using our interface (\cref{sec:retrieval_model}) and (2) information synthesis (\cref{sec:synthesis_model}). The overall framework is illustrated in Figure~\ref{fig:method}.

\subsection{Search Model}
\label{sec:retrieval_model}
\paragraph{Overview.} We partition web search into $3$ sub-tasks: action prediction, search query generation, and supporting fact extraction. Each task is cast as a text-to-text format and we train $3$ separate modules using a generative PLM. By combining the $3$ modules, we build the search model, which executes a series of actions to gather relevant information. The action prediction module decides which action to perform at each step. If the module predicts \texttt{Search} or \texttt{Quote} as the current action, then it calls the other two modules to generate the contents of the query or the supporting fact.

Each module performs inference conditioned on the current state $\mathcal{S}_t$ of the interface at time step $t$. $\mathcal{S}_t$ comprises the original question $\mathcal{Q}_0$, the query currently searching $\mathcal{Q}_t$, the past action sequence $\mathcal{A}_{t-1} \!=\! \{a_1, ..., a_{t-1}\}$, the last and the current content displayed in the window $\mathcal{W}_{t-1}$ and $\mathcal{W}_t$, current supporting facts $\mathcal{F}_t \!=\! \{f_1, ..., f_{|\mathcal{F}_t|}\}$, and the number of remaining actions. If an action is executed, the components of $\mathcal{S}_t$ would be updated. 
$\mathcal{W}$ can be either the three links in the \textit{search mode} or the specific page content in the \textit{browsing mode}. We only maintain the recent two observations ($\mathcal{W}_{t-1}$ and $\mathcal{W}_t$) displayed in the window instead of concatenating all the past observations because the latter may exceed the input length limit of the PLM. Next, we introduce the three modules in detail.

\paragraph{Action Prediction.} This module predicts which action to perform next. Since there are $10$ possible actions in total, action prediction can be viewed as a $10$-category classification task. Take the action \texttt{Search} as an example, denote $\{x_1, ..., x_{\text{N}}\}$ as the tokenized sequence for the action name \texttt{Search}, where $x_*$ denotes a specific token. The probability of \texttt{Search} can be factorized as follows:
\begin{equation}
\small
\begin{aligned}
    \mathcal{P}(\texttt{Search}|\mathcal{S}_t) = \mathcal{P}(x_1|\mathcal{S}_t) \times \prod_{i=2}^{\text{N}}\mathcal{P}(x_i|\mathcal{S}_t,x_1, ..., x_{i-1}).
    \nonumber
\end{aligned}
\end{equation}
During inference, we select the action with the highest probability to perform on the interface.

\paragraph{Search Query Generation.} This module generates a query $\mathcal{Q}_{t+1} \!=\! \{q_1, ..., q_{|\mathcal{Q}_{t+1}|}\}$ to search Bing, which is also formulated as text generation:
\begin{equation}
\small
\begin{aligned}
    \mathcal{P}(\mathcal{Q}_{t+1}|\mathcal{S}_t) = \mathcal{P}(q_1|\mathcal{S}_t) \!\times\! \prod_{i=2}^{|\mathcal{Q}_{t+1}|}\mathcal{P}(q_i|\mathcal{S}_t,q_1, ..., q_{i-1}).
    \nonumber
\end{aligned}
\end{equation}

\paragraph{Supporting Fact Extraction.} Assume in the \textit{browsing mode}, the current content of the window is $\mathcal{W}_t \!=\! \{w_1, ..., w_{|\mathcal{W}_t|}\}$. We aim to extract a supporting fact $f \!=\! \{w_i, ..., w_j\}$ from $\mathcal{W}_t$, where $1 \!\le\! i \!\le\! j \!\le\! |\mathcal{W}_t|$. While a naive solution is to directly generate all the tokens of $f$ auto-regressively, this solution suffers from low inference speed in practice. As an alternative, we only generate the first and last few ($\text{N}_f$) tokens of $f$ given $\mathcal{S}_t$. Formally, we maximize $\mathcal{P}(\texttt{[s]}, w_i, ..., w_{i\text{+N}_f\text{-1}}, \texttt{[e]}, w_{j\text{-N}_f\text{+1}}, ..., w_{j}|\mathcal{S}_t)$,
where \texttt{[s]} and \texttt{[e]} denote the special tokens that indicate the start and end of the fact $f$. During inference, after decoding the start and end tokens, we can locate the desired sequence in $\mathcal{W}_t$ by text matching. If the start / end tokens occur in multiple locations of $\mathcal{W}_t$, we always extract the longest sequence from $\mathcal{W}_t$, and a large $\text{N}_f$ could lower the frequency of this multi-location issue. Note disjoint spans in $\mathcal{W}_t$ can be extracted by executing multiple \texttt{Quote} actions consecutively.

\subsection{Synthesis Model}
\label{sec:synthesis_model}
The \textbf{information synthesis} task learns to organize a series of supporting facts into a coherent answer. However, not as perfect as humans, the trained search model occasionally gathers irrelevant noises, which would influence the quality of the generated answer. To remedy this, we corrupt the collected facts in the training data of the synthesis model by introducing noises. Specifically, given a series of human-extracted facts $\{f_1, ..., f_{\text{N}}\}$, we randomly select a few unrelated facts $\{f'_1, ..., f'_{\text{N}'}\}$ from other training instances. After randomly shuffling all the facts, we concatenate them as the final input. During training, the model is optimized to generate the human-annotated answer conditioned on the corrupted supporting facts, i.e., maximizing $\mathcal{P}(Answer|\mathcal{Q}_0, f_1, ..., f_{\text{N}}, f'_{1}, ..., f'_{\text{N}'})$.
Since the annotated answer does not contain the information of $f'_*$, the model learns to ignore irrelevant facts and only focus on important ones for generation.
\section{Experiments and Analyses}
\label{sec:main_exp}
Our problem consists of $4$ sub-tasks: action prediction, search query generation, supporting fact extraction, and information synthesis. Correspondingly, we first train $4$ modules and evaluate each sub-task independently by feeding the ground truth input to each module (\cref{sec:exp_sub_task}). Then we combine all modules into a unitary pipeline and only feed the question to the pipeline for a holistic evaluation (\cref{sec:exp_pipeline}). Finally, we conduct in-depth analyses for each module to understand their behaviors (\cref{sec:in_depth_analysis}).

\subsection{Individual Sub-task Evaluation}
\label{sec:exp_sub_task}
\paragraph{Settings.}
We evaluate $8$ typical generative PLMs that support Chinese, covering $3$ architectures:
\begin{itemize} [topsep=1pt, partopsep=1pt, leftmargin=12pt, itemsep=-3pt]
    \item $\text{T5}$ architecture~\citep{raffel2019exploring}: $\textbf{mT5}_\textbf{BASE}$~\citep{xue-etal-2021-mt5}, a $580$M model pre-trained on mC4; $\textbf{mT0}_{\textbf{BASE}}$~\citep{muennighoff2022crosslingual}, which fine-tunes $\textbf{mT5}_\textbf{BASE}$ on diverse downstream tasks; $\textbf{Mengzi-T5}_\textbf{BASE}$~\citep{zhang2021mengzi}, a $220$M model pre-trained on $300$G internet corpora.
    \item BART architecture~\citep{lewis-etal-2020-bart}: $\textbf{mBART}_{\textbf{LARGE}}$~\citep{liu-etal-2020-multilingual-denoising}, a $680$M model pre-trained on monolingual corpora of multiple languages; $\textbf{C-BART}_\textbf{LARGE}$~\citep{shao2021cpt}, a $406$M model pre-trained on $200$G web texts.
    \item CPM architecture~\citep{zhang2021cpm}: $\textbf{CPM}_\textbf{2.6B}$, $\textbf{CPM}_\textbf{7B}$, and $\textbf{CPM}_\textbf{10B}$, which contain $2.6$B, $7$B, and $10$B parameters, respectively, and are pre-trained with increasing sizes of data.
\end{itemize}

Among these PLMs, $\textbf{mT5}_\textbf{BASE}$, $\textbf{mT0}_\textbf{BASE}$, and $\textbf{mBART}_{\textbf{LARGE}}$ are multilingual and the others are Chinese-only PLMs. We elaborate on details of the above PLMs in \cref{sec:PLM_detail}. We adopt recommended fine-tuning configurations of the original papers for all PLMs.
For evaluation metrics, we treat action prediction as a $10$-category classification task and choose \textit{Micro-F1} and \textit{Macro-F1} as the metric. We treat the other three tasks as text generation and calculate \textit{Rouge-L} of the generated sequence and the ground truth.

\paragraph{Results.}
The results are listed in Table~\ref{tab:individual_eval}, from which we conclude that: (1) $\textbf{mT0}_{\textbf{BASE}}$ outperforms $\textbf{mT5}_{\textbf{BASE}}$ in action prediction, query generation, and supporting fact extraction, but performs poorer in information synthesis. We conjecture this is because $\textbf{mT0}_{\textbf{BASE}}$ enhances language skills more related to the first three tasks during its multi-task fine-tuning. Rather, the information synthesis ability might have been weakened. Besides, $\textbf{Mengzi-T5}_\textbf{BASE}$ performs generally well on all tasks despite owning much fewer parameters; (2) in general, $\textbf{mBART}_{\textbf{LARGE}}$ and $\textbf{C-BART}_\textbf{LARGE}$ show inferior performance than all other PLMs, except that $\textbf{mBART}_{\textbf{LARGE}}$ exhibits excellent performance in information synthesis; (3) comparing the results of $\textbf{CPM}_\textbf{2.6B}$, $\textbf{CPM}_\textbf{7B}$, and $\textbf{CPM}_\textbf{10B}$, we find that \textbf{the performance generally gets improved as the model size increases}. Blessed by the scaling law~\citep{kaplan2020scaling}, larger PLMs own stronger understanding and generation abilities and could achieve better downstream performance.

\begin{table}[!t]
  \centering
  \small
    \begin{tabular}{@{~}c@{~~}c@{}c@{}c@{}c@{}c@{}}
    \toprule
    \textbf{Task}  & \multicolumn{2}{c}{\textbf{Action}} & \multicolumn{1}{c}{\textbf{Query}} & \multicolumn{1}{c}{\textbf{Fact}} & \multicolumn{1}{c}{\textbf{Synth.}} \\
    \textbf{Metric} & \multicolumn{1}{c}{\textit{Mi.}} & \multicolumn{1}{c}{\textit{Ma.}} & \multicolumn{1}{c}{\textit{R-L}} & \multicolumn{1}{c}{\textit{R-L}} & \multicolumn{1}{c}{\textit{R-L}} \\
    \midrule
    $\textbf{mT5}_\textbf{BASE}$   & $53.8$ & $44.0$ & $62.4$ & $56.7$ & $56.8$ \\
    $\textbf{mT0}_\textbf{BASE}$ & $58.2$ & $52.1$ & $64.6$ & $60.0$ & $51.4$ \\
    $\textbf{Mengzi-T5}_\textbf{BASE}$ & $58.1$ & $51.2$ & $62.6$ & $61.9$ & $57.7$ \\
    $\textbf{mBART}_\textbf{LARGE}$ & $53.6$ & $41.1$ & $50.4$ & $56.5$ & $60.2$ \\
    $\textbf{C-BART}_\textbf{LARGE}$ & $43.8$ & $31.3$ & $56.1$ & $49.3$ & $50.6$ \\
    $\textbf{CPM}_\textbf{2.6B}$ &   $55.6$    &  $49.8$     &   $61.6$    &  $52.6$     &  $55.0$ \\
    $\textbf{CPM}_\textbf{7B}$ &  $58.9$     &  $50.5$     &   $67.8$    &   $59.8$    &  $56.4$ \\
    $\textbf{CPM}_\textbf{10B}$ & $\textbf{60.4}$ & $\textbf{54.5}$ & $\textbf{70.0}$   &   \textbf{62.4}    &  $\textbf{61.2}$ \\
    \bottomrule
    \end{tabular}%
  \caption{Sub-task evaluation (test performance) using $8$ PLMs. We report \textit{Micro-F1} (\textit{Mi.}), \textit{Macro-F1} (\textit{Ma.}) for action prediction, and \textit{Rouge-L} (\textit{R-L}) for query generation, fact extraction, and information synthesis.}
  \label{tab:individual_eval}%
\end{table}%

\begin{figure*}[!t]
    \centering
    \subfigure{\includegraphics[width=0.95\textwidth]{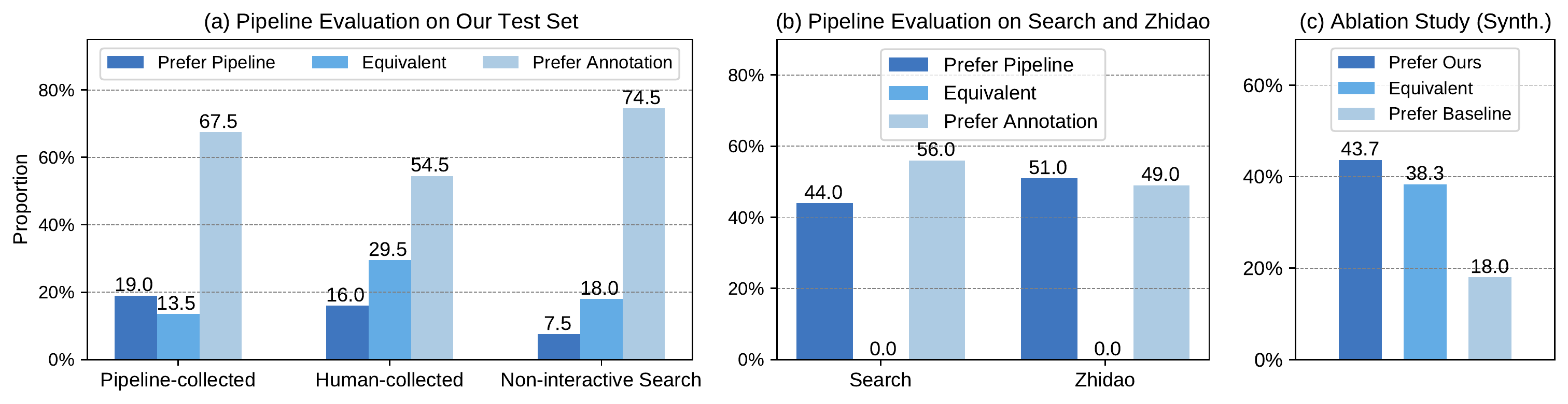}} 
    \caption{Results of human evaluation. (a) Comparison of human annotation and our pipeline-generated answers with different sources of supporting facts. (b) The experiments on two DuReader datasets: Search and Zhidao. We compare our pipeline with the golden annotation. (c) Ablation study for our synthesis model.}
    \label{fig:pipeline_compare}
\end{figure*}

\subsection{Holistic Pipeline Evaluation}
\label{sec:exp_pipeline}
We choose the modules trained by $\textbf{CPM}_\textbf{10B}$, which performs the best among all the PLMs in \cref{sec:exp_sub_task}, and combine them into the overall pipeline. Then we evaluate its performance compared with humans.

\paragraph{Compared Answer Pairs.}
For each test question of \ourmodel, we compare the annotated answer with $3$ types of answers generated by our synthesis model. Specifically, the $3$ types of answers differ in the source of supporting facts, including (1) the facts collected by our search model, (2) ground-truth human-collected facts, and (3) the facts collected using a commonly adopted non-interactive web search method. For (3), we directly input the original question into Bing, extract the paragraphs from all the retrieved links, and rank them using TF-IDF. Then we concatenate the top-$k$ paragraphs as the input until it exceeds $3072$ tokens.

\paragraph{Evaluation Protocol.}
We engage $8$ annotators to manually compare different answers based on human preference. Given a question and a pair of answers, we ask them to perform an overall assessment and decide which answer they would prefer based on multiple factors, including the overall usefulness, coherence, and relevance to the question. Since all three retrieval methods use the same search engine, their collected facts sometimes have a high overlap, which leads to similar answers. Thus we allow annotators to mark two answers as \textit{equivalent} if both are of comparable quality.

\paragraph{Results.}
We derive from the results in Figure~\ref{fig:pipeline_compare} (a) that: (1) the answers obtained purely by our pipeline are preferred or comparable to human-written answers $19.0\% \!+\! 13.5\% \!=\! 32.5\%$ of the time. This result implies ample opportunity for advancement of our pipeline in future endeavors, which is discussed in \cref{sec:future_work}. (2) When applying our synthesis model to the human-collected facts, the performance grows to $16.0\% \!+\! 29.5\% \!=\! 45.5\%$ preference or equivalence, which is due to the improved quality of the collected facts. (3) The facts gathered by non-interactive search lead to slightly worse performance ($7.5\% \!+\! 18\% \!=\! 25.5\%$) than our search model. The \textbf{superiority of our search model over non-interactive search} may be because our model (a) sends diverse queries to Bing multiple times so that more abundant information can be retrieved, and (b) it critically decides whether a web page contains important information, which performs better than TF-IDF.

\paragraph{Experiments on DuReader.}
Next, we apply our pipeline (search model and synthesis model) to $2$ Chinese QA datasets from DuReader, i.e., Zhidao and Search. Although not specially designed for LFQA, DuReader contains a variety of question types, and we randomly sample $400$ test questions whose annotated answers are longer than $400$ Chinese characters. For these questions, we engage annotators to compare our pipeline-generated answers with the golden annotations of DuReader. From the results in Figure~\ref{fig:pipeline_compare} (b), we find that our pipeline generates answers better than the annotated ones $44.0\%$ and $51.0\%$ of the time on Search and Zhidao ($47.5\%$ on average), showing satisfying out-of-distribution generalization performance. The fact that the same pipeline surpasses fewer human-written answers on our dataset than DuReader also reflects \textbf{the high quality of our annotated answers}. Note the \textit{equivalent} ratio is $0\%$ because both answers are based on totally different supporting facts, and it is easy to determine which one is better.

\subsection{Further Analysis}
\label{sec:in_depth_analysis}
Next, we conduct in-depth analyses to gain a deeper understanding of each module. Without loss of generality, we evaluate $\textbf{CPM}_\textbf{7B}$ in this section.

\paragraph{Ablation Study for the Synthesis Model.}
We evaluate whether corrupting the synthesis model's training data by introducing irrelevant facts improves its ability to ignore noisy facts. We train a baseline model without corrupting the training data and keep other settings the same as our model. For each test question, we feed the supporting facts collected by our search model to both synthesis models and generate two answers. Annotators would evaluate which answer is more relevant to the original question (the \textit{equivalent} option is allowed).

According to Figure~\ref{fig:pipeline_compare} (c), by corrupting the training data, our model performs better than the baseline $43.7\%$ of the time and is worse $18.0\%$ of the cases. This demonstrates that \textbf{our method indeed enhances the model's ability to ignore noisy information}, which makes the generated answer more relevant to the original question. In \cref{sec:novel_generation}, we further explore the use of another corruption method that flexibly balances generating novel contents and copying supporting facts.

\begin{table}[!t]
  \centering
  \small
    \begin{tabular}{c@{~~}c@{~~~}c@{~~~}c@{~~~}c@{~~}c@{~~}}
    \toprule
    \textbf{Task}  & \multicolumn{2}{c@{~~~~~~}}{\textbf{Action}} & \textbf{Fact} &   \textbf{Task}    & \textbf{Query} \\
    \textbf{Metric} & \textit{Mi.} & \textit{Ma.} & \textit{R-L} &  \textbf{Metric}      & \textit{R-L} \\
    \midrule
    $\mathcal{S}_t$  &  $58.9$     &   $50.5$    &  $59.8$   & $\mathcal{S}_t$  &  $67.8$ \\
    - $\mathcal{F}_t$ &   $55.5$    &  $49.3$     & $54.7$ & - $\mathcal{F}_t$ &  $66.9$ \\
    - $\mathcal{W}_{t-1}$ &   $57.7$    &  $52.0$     &   $59.3$     & - past queries $\!\in\! \mathcal{A}_{t-1}$ &  $65.3$ \\
    - $\mathcal{A}_{t-1}$ &   $53.4$    &   $44.1$    &  $60.3$     & - seen titles $\!\in\! \mathcal{A}_{t-1}$ &  $65.3$ \\
    \bottomrule
    \end{tabular}%
  \caption{Ablation study of the search model when different components are removed from $\mathcal{S}_t$, respectively.}
  \label{tab:ablation_state}%
\end{table}%

\paragraph{Effects of Components in $\mathcal{S}_t$.}
We conduct ablation studies for several components of $\mathcal{S}_t$ to examine how they contribute to each module of the search model. This is achieved by modifying both the training and evaluation data of each module. For action prediction and supporting fact extraction, we remove one of the following: the existing collected facts $\mathcal{F}_t$, the contents displayed in the last window $\mathcal{W}_{t-1}$, or the past actions $\mathcal{A}_{t-1}$. For query generation, the following items are removed from $\mathcal{S}_t$: the existing collected facts $\mathcal{F}_t$, the already searched queries, or the titles of the links browsed before. The information of the latter two items is included in $\mathcal{A}_{t-1}$. Specifically, for the past action \texttt{Search} / \texttt{Load Page}, $\mathcal{A}_{t-1}$ not only includes the action name, but also records the specific searched query / the title of the loaded page.

The results are listed in Table~\ref{tab:ablation_state}, from which we observe that: (1) for action prediction, the removal of either $\mathcal{F}_t$ or $\mathcal{W}_{t-1}$ only leads to minimal performance changes, while removing $\mathcal{A}_{t-1}$ leads to a significant performance drop. This shows that \textbf{the past actions are critical factors for action prediction}; (2) for supporting fact extraction, only removing $\mathcal{F}_t$ impairs the performance significantly ($-5.1$). This indicates that aligned with humans, \textbf{the module considers what has been extracted to decide which information to extract next}; (3) for query generation, removing either searched queries or accessed link titles in $\mathcal{A}_{t-1}$ causes a great negative impact ($-2.5$), which means \textbf{the module might have learned to generate queries based on what has been searched and newly observed information during web search}. This feature is humanoid in that humans also consider both information to avoid sending repetitive queries and to ask follow-up questions about an accessed link.

\begin{figure}[!t]
    \centering
    \subfigure{\includegraphics[width=0.45\textwidth]{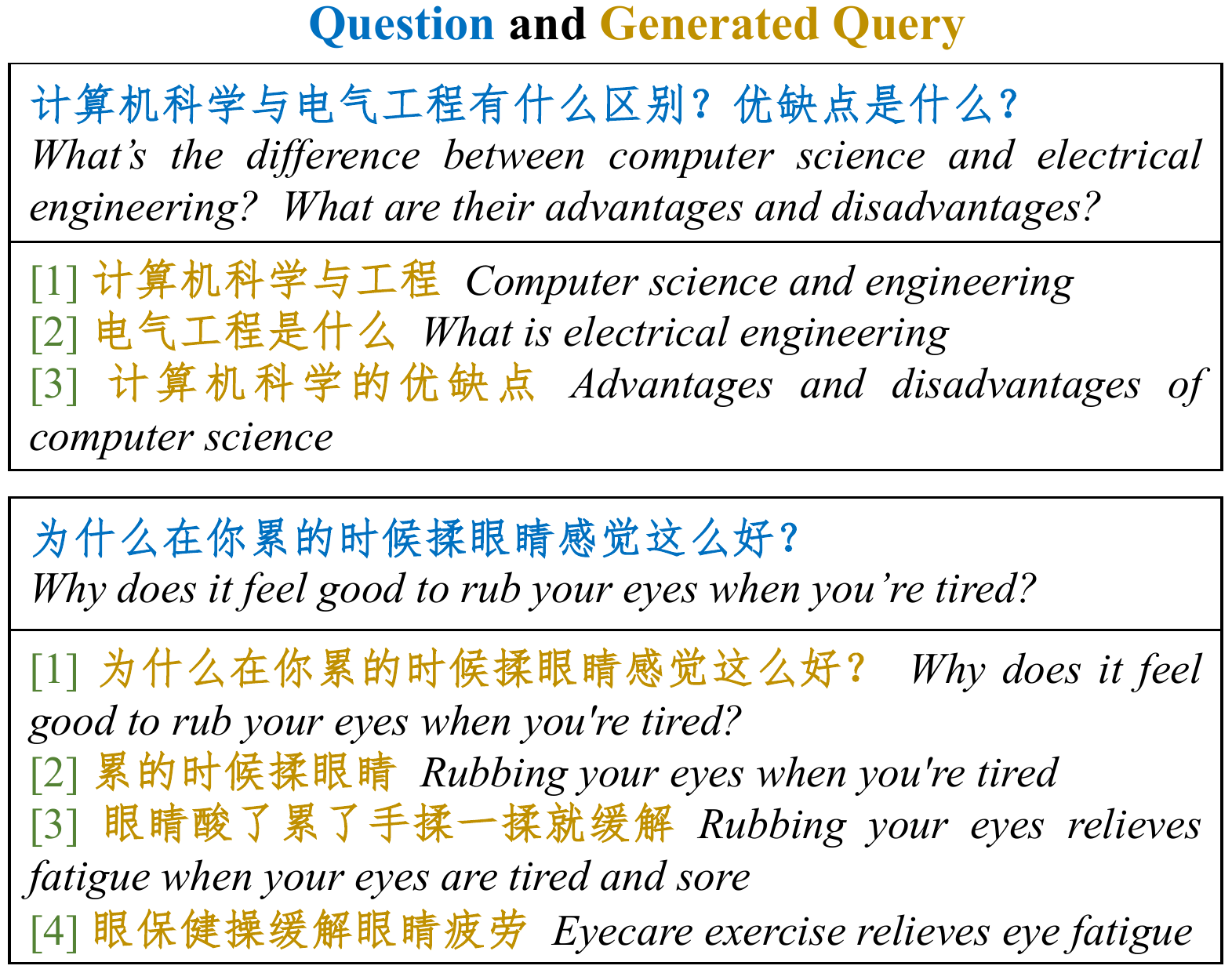}}
    \caption{Case study for query generation. We show the generated queries for two test questions.}
    \label{fig:query_case_study}
\end{figure}

\paragraph{Case Study for Query Generation.}
To fathom the human behaviors learned by our query module, we conduct a case study by sampling the generated queries for different questions in the test set. We illustrate two representative results in Figure~\ref{fig:query_case_study} to showcase the typical strategies learned by our query module, including copying the original question, decomposing the question into multiple sub-questions, rephrasing questions with related terms, etc. These strategies make the queries more diverse, which helps gather more abundant information from various sources.
\section{Conclusion}
In this paper, we construct a benchmark of interactive web search for Chinese long-form QA, together with an open-source interface. We decompose the task into $4$ sub-tasks and design a modular pipeline. By fine-tuning representative PLMs, we conduct both an individual evaluation for each module and a holistic evaluation for the pipeline. In-depth analyses are carried out to understand the core design elements of our framework. We expect our interface, dataset, framework, and analyses to facilitate more future explorations in this area.

\section*{Acknowledgments}

This work is supported by the National Key R\&D Program of China (No. 2020AAA0106502), Institute Guo Qiang at Tsinghua University, Beijing Academy of Artificial Intelligence (BAAI). Huadong Wang is funded by China Postdoctoral Science Foundation (No. 2022M721829).

Yujia Qin and Zihan Cai led the data collection. Yujia Qin, Dian Jin, Lan Yan, Shihao Liang, and Kunlun Zhu conducted the experiments. Yujia Qin wrote the paper. Yankai Lin, Xu Han, Zhiyuan Liu, Maosong Sun, and Jie Zhou advised the project. All authors participated in the discussion. The authors would like to thank Aran for the implementation of the interface, Shengding Hu, Haoyang Pang and Zhenhuan Huang for the discussion, and the anonymous annotators for their huge efforts.

\section*{Limitations}

The human evaluation shows that our pipeline performs worse than humans in the process of information retrieval and synthesis $67.5\%$ of the time, which still leaves room for improvement (see \cref{sec:future_work} for future works).

\section*{Ethical Statement}
In this research, we adhere to the highest ethical standards and commit to making every effort to minimize any potential harm. Specifically:

\begin{itemize} [topsep=1pt, partopsep=1pt, leftmargin=12pt, itemsep=-3pt]
    \item When creating our dataset, we have ensured that all data collected is obtained through legitimate and legal means. In addition, we have obtained the appropriate permissions and consent from all necessary parties.
    \item We have also taken steps to protect the privacy of individuals whose data is included in our dataset through de-identification during annotation.
    \item We are committed to eliminating bias, discrimination, or stereotypes during annotation by removing any suspect examples.
    \item We take the responsibility of open-sourcing the interface, dataset, codes, and trained models to the public. However, there are cases that these resources are maliciously used. For instance, our models may be utilized to generate responses without proper attribution of the information source, causing severe consequences. We would strive to ensure that they are used ethically and not for any malicious or harm-causing intent.
\end{itemize}
\bibliography{anthology,custom}
\bibliographystyle{acl_natbib}

\clearpage
\appendix

\section*{Appendices}
\label{sec:appendix}

\section{Implementation Details of the Interface}
\label{sec:interface_detail}
Our interface includes two components: an API back end and a website front end.

\subsection{API Back End}
\label{sec:API_back_end}
The API backend implements three APIs with different functions: (1) \textit{search}, which receives queries from users and returns search results recommended by Bing; (2) \textit{extract}, which receives a URL and returns the text-only contents of the corresponding web page; (3) \textit{record}, which receives the actions conducted by agents and stores them in a database. 

\paragraph{Search API.}
The search API is based on Bing API. When it receives keywords from users, it calls Bing API to search for relevant results and converts them into the format we specify. Each result consists of a title, the link to the page, and a brief summary of the page contents. To ensure the originality of the answers generated during annotation, we have implemented a filter in the search API to exclude results from certain websites (e.g., Reddit forums). This is necessary because some of the questions are sourced from websites that may appear in search results.

\paragraph{Extract API.}
The contents of web pages often include huge quantities of layout information and multimedia that is inappropriate to display directly to agents and is meaningless for our task. Therefore, we use a third-party tool\footnote{\url{https://github.com/mozilla/readability}} to extract the simplified text-only contents of web pages. This ensures that only clean and meaningful text will be presented to the users.

\paragraph{Record API.}
Actions conducted by users are recorded in the website front end, when users finish the annotation process of a question, the front end will call this Record API, and the detailed action information and meaningful observations during web search will be uploaded and stored in our database.

\subsection{Website Front End}
\label{sec:website_front_end}
The website front end is designed as a graphic user interface for human annotators, which supports two modes: the \textit{search mode} and the \textit{browsing mode}. Each time an action is performed, it will be recorded and the corresponding changes will be rendered in our website and displayed to the users. 

\begin{figure*}[!t]
    \centering
    \subfigure{\includegraphics[width=0.7\textwidth]{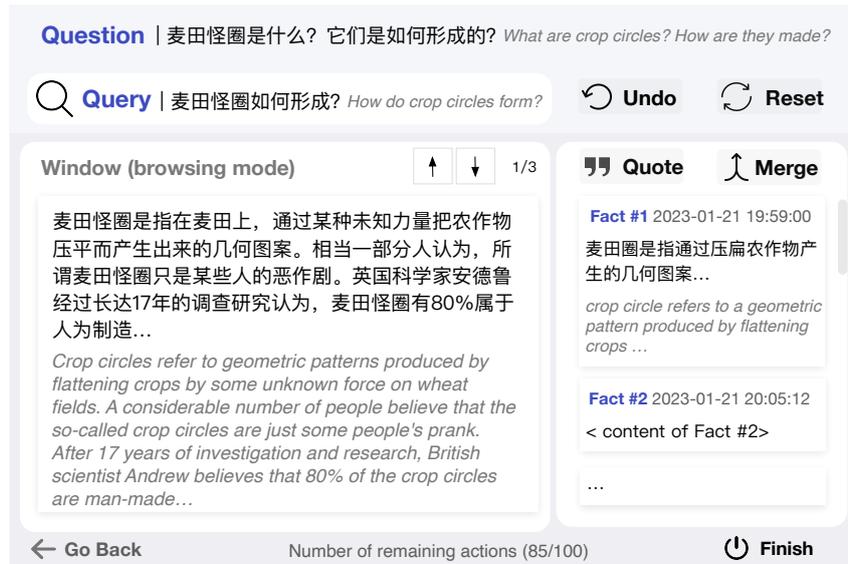}}
    \caption{An example of \textit{browsing mode} of our interface.}
    \label{fig:browsing_mode}
\end{figure*}

\paragraph{Window.}
In the \textit{search mode}, the window displays the searched results returned by our API back end. We present at most three links at a time in each window, and the \texttt{Scroll} action can be used to access other links. In the \textit{browsing mode}, when clicking a specific link, \texttt{Load Page} action is triggered and the front end will call the extract API and display the text-only contents of the web page. The length of content in each window is limited up to $500$ Chinese characters, and the \texttt{Scroll} action can be used to access more content. In the main paper, we illustrate an example for the \textit{search mode} of our interface, here we present the example for the \textit{browsing mode} in Figure~\ref{fig:browsing_mode}. In addition, we also display the existing supporting facts and the remaining number of actions for ease of human annotation.

\paragraph{Actions.}
Once an action is performed, we record the current state of the interface, which includes the content displayed in the window, the current query issued, the existing collected supporting facts, the remaining number of actions, etc. We also record the specific information about the current action, for instance, \texttt{Search <query>} includes the content of the query, \texttt{Load Page <idx>} includes all the detailed information about a web page, and \texttt{Quote <content>} includes the consecutive sentences selected by the user.

It should be noted that the action \texttt{Merge} is specially designed for extracting a supporting fact that crosses the boundary of two windows in the \textit{browsing mode}. For instance, the user can perform \texttt{Quote <content1>}, \texttt{Scroll Down}, \texttt{Quote <content2>}, and \texttt{Merge} to get one supporting fact, which is concatenated by both \texttt{content1} and \texttt{content2}.

Besides, we also implement (1) the \texttt{Undo} action, which supports revoking the last action performed, and (2) the \texttt{Reset} action, which terminates the current annotation and starts a new one. Both actions will not be recorded since they do not belong to meaningful web search behaviors.

\section{Annotation Principle}
\label{sec:annotation_principle}
Below we present the annotation principles for web search, supporting fact extraction, and question answering. These principles are part of our annotation guides, which are sent to our contractors before annotation. The original version of the following is written in Chinese, and we have translated it into English.

\subsection{Web Search Principle}

\paragraph{Look for Relevant Information.}
In the search process, it is important to ensure that the content being searched is closely related to the question at hand. During the labeling process, users may encounter various concepts that are related to the question but may not be central to the main idea. These peripheral concepts should be ignored in the search. For instance, when searching for information about ``the principle of the constant speed of light'', it is possible to come across the concept of ``Lorentz transformation'', which is related to the topic but only tangentially. As such, it is not necessary to include a detailed explanation of ``Lorentz transformation''.

\paragraph{Send Simple Queries.}
Search engines are often less effective when the question being asked is long and complex. In such cases, it is advisable to simplify and refine the main question or keywords to improve the chances of finding relevant information and reduce the number of unnecessary search actions. For example, instead of searching for the question ``I have a question that bothers me a lot, why do most crustaceans / seafood turn from light gray to red / orange when heated?'', it would be more effective to simplify it to ``why does seafood change color when heated?''. This ensures the simplicity of the queries, making it more likely to find relevant information.

\paragraph{Avoid Unnecessary Search.}
Search engines typically rank web pages based on their relevance to the query, with higher-ranked results being more relevant. If the top-ranked results for a particular search do not align with the user's needs, it may not be productive to continue scrolling through the results to find relevant information. Instead, it is more efficient to issue a new query to reduce the number of unnecessary search actions.

\subsection{Supporting Fact Extraction Principle}

\paragraph{Find Diverse Relevant Facts.}
The supporting facts should contain information that is relevant to the original question. When possible, it is generally more effective to extract supporting facts from diverse sources, while ensuring that the content remains highly relevant to the original question. It is important to avoid duplicating summaries of the same content from different sources, as this does not contribute to answering the question.

\paragraph{Avoid Recording Fragmentary Facts.}
The extracted supporting fact should contain complete and coherent information. It is important to avoid intercepting sentences with incomplete semantics or taking them out of context, as this can alter the meaning of the supporting fact. In addition, please ensure the integrity of the supporting fact by including all relevant information and expressing it in a coherent manner.

\paragraph{Ensure the Factual Accuracy.}
It is important to summarize information from trusted sources whenever possible. This helps ensure the reliability of the information being used. You can also judge the factual accuracy of a supporting fact by comparing it with other searched results.

\subsection{Answer Principle}
A good long-form answer is typically well-researched, well-written, and provides a thorough and detailed response. It should be well-organized and easy to read, with clear and concise language that is appropriate for the intended audience. Additionally, a good answer should be objective and unbiased, presenting multiple viewpoints on the topic if applicable.

\paragraph{Coherence and Relevance.}
Coherence refers to the overall logical consistency and clarity of the answer. The desired answer should have a clear structure, with each paragraph building upon the previous one and contributing to the overall argument. The ideas presented should flow smoothly and be easy to follow. Relevance means the extent to which the answer addresses the original question. The desired answer should stay on topic, providing information that is directly relevant to the question. It should not include unnecessary or tangential information. Together, coherence and relevance help guarantee that the answer is easy to understand and stays focused on the main topic, making it more useful and informative for the reader.

\paragraph{Objectivity.}
The content of the answer should be based on the information obtained during the search process. The desired answer should present information and viewpoints in an unbiased manner, without expressing personal opinions or preferences. While the annotation process inevitably involves subjectivity, the questions are relatively straightforward and it should not be difficult to maintain a degree of objectivity. Please be neutral and fair, and present multiple sides of an issue if applicable.

\paragraph{Conciseness.}
There is no specific word count requirement for answers, but it is important to provide concise, comprehensive, and in-depth answers that include necessary auxiliary information. It is generally best to avoid extremely long or short answers. In addition, the sentences in the answer should be concise and clear and should avoid redundancy. For example, the question ``How toxic is barium chloride?'' should not be answered simply with ``very toxic''. Instead, a more detailed description of the toxicity of barium chloride, including the poisoning dose, poisoning symptoms, and poisoning mechanism, would be more informative and useful. It is important to provide a well-rounded and thorough answer to the question, rather than just a brief or overly general response.

\paragraph{Normative.}
It is important to answer questions in written language, as this can help make the answer more formal. Annotators should avoid using irregular or unconventional expressions that may not be understood by everyone. Typos or grammatical errors are not allowed.

\section{More Details for Data Collection}
\label{sec:data_collection_detail}

We limit our annotators and quality inspectors to native Chinese speakers. We make sure all our annotators are fairly compensated by the market price.

\paragraph{Question Creation.} 
Chinese QA forums, such as \href{https://www.zhihu.com/}{Zhihu} and \href{https://zhidao.baidu.com/}{Baidu Zhidao},
are known for their abundance of long-form questions. However, when these questions are utilized as direct queries on Bing, users can often access multiple websites that contain well-organized answers, thus making the web search process less challenging. Such an issue is not mitigated even if we block the source from \href{https://www.zhihu.com/}{Zhihu} and \href{https://zhidao.baidu.com/}{Baidu Zhidao}. In view of this, we strive to annotate new open-ended questions that have not been answered on Chinese QA forums.

Following \textsc{ELI5}~\citep{fan-etal-2019-eli5}, we turn to creating questions from Reddit forums\footnote{\url{https://www.reddit.com/r/explainlikeimfive}} as an alternative. We closely follow the way \textsc{ELI5} collects the source questions. After collection, we engage annotators to refer to these questions and then ask new questions in Chinese. This way significantly improves the productivity of question creation.

For quality control, our quality inspectors would check whether the created question is meaningful, semantically coherent, comprehensible, and reasonable. Only those questions that satisfy the above requirements would be retained. In addition, we also remove the questions that are politically sensitive. In total, $22.4\%$ newly created questions are discarded.

\paragraph{Web Search and Answer Annotation.}
Before annotation, we provide our annotators with detailed annotation guidance. They got paid based on the number of instances they annotate instead of the time spent during annotation. Note for answer annotation, we did not require annotators to use all the collected facts when composing the answer but asked them to record which facts are leveraged in their answer.

\begin{figure}[!t]
    \centering
    \subfigure{\includegraphics[width=0.45\textwidth]{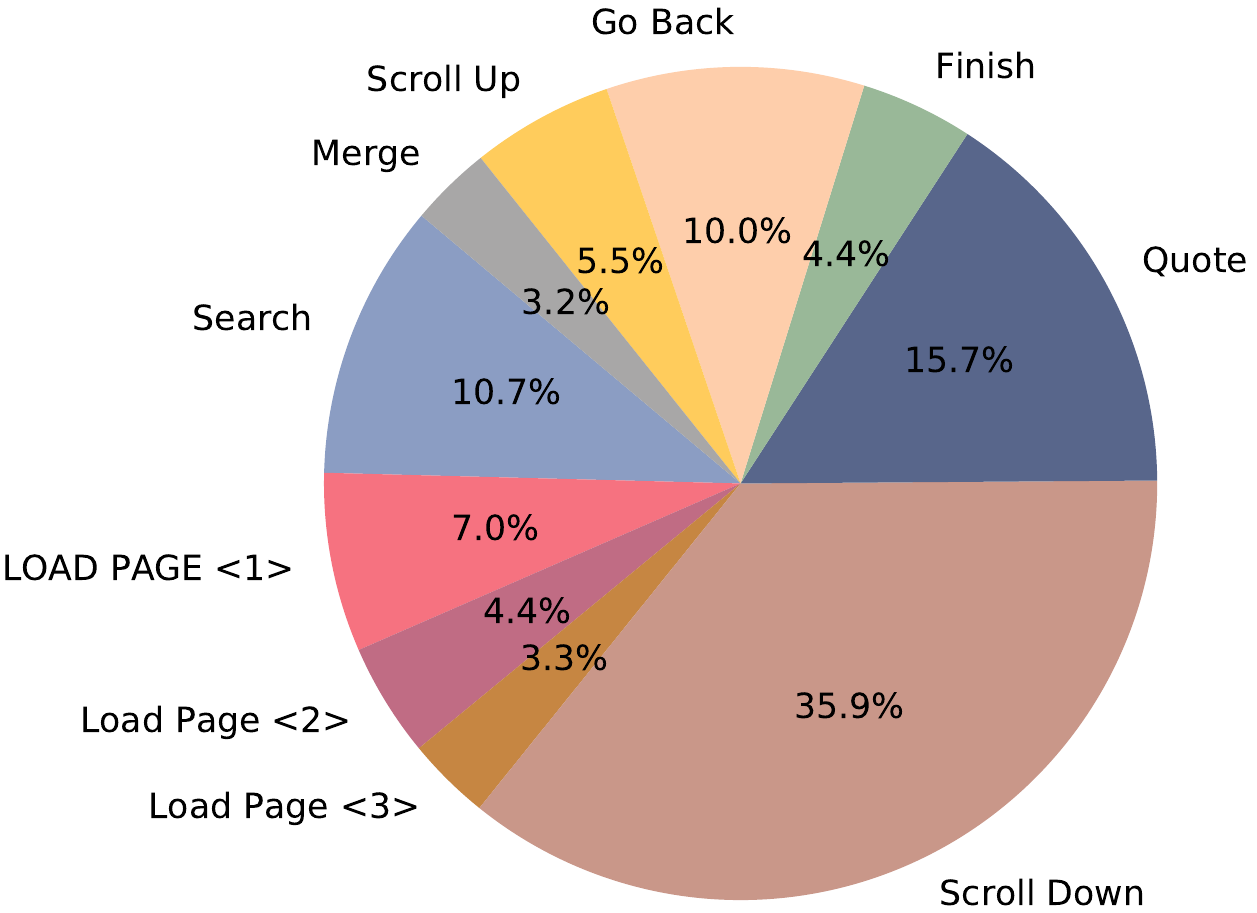}}
    \caption{Proportion of different pre-defined actions in our \ourmodel.}
    \label{fig:action_pie}
\end{figure}

\paragraph{Proportion for Different Actions.}
We record the proportion of different pre-defined actions in our collected dataset in Figure~\ref{fig:action_pie}. As can be seen, \texttt{Scroll Down}, \texttt{Quote}, and \texttt{Search} are the most frequently used actions. The proportion of \texttt{Load Page <1>} is larger than those of \texttt{Load Page <2>} and \texttt{Load Page <3>}. This is because search engines rank search results based on their relevance to the query. Humans tend to visit the links according to the order recommended by search engines. If humans have collected enough supporting facts on the first page or find it to be irrelevant, they probably would not continue browsing other web pages of the current query.

\section{Details for the PLMs Evaluated}
\label{sec:PLM_detail}
We select $6$ series of representative and publicly available generative PLMs that support Chinese. For all the models, we use them for their intended uses. In the following, we give a brief introduction to them:

$\textbf{mT5}$~\citep{xue-etal-2021-mt5} is a multilingual encoder-decoder PLM with a general-purpose text-to-text format. Its pre-training data mC4~\citep{xue-etal-2021-mt5} covers $101$ languages collected from the public Common Crawl web scrape. $\textbf{mT5}$ achieves superior performance in various multilingual benchmarks.

\textbf{mT0}~\citep{muennighoff2022crosslingual} is a multi-task fine-tuned version of Google's \textbf{mT5}. The model attained strong zero-shot performance and cross-lingual generalization ability. Through explicit multi-task learning, a variety of language capabilities are enhanced through knowledge transfer; inevitably, some capabilities, which are not required by the trained tasks, might have been impaired.

\textbf{Mengzi-T5}~\citep{zhang2021mengzi} is a powerful Chinese encoder-decoder PLM that achieved state-of-the-art results on the CLUE benchmark. Instead of chasing a larger scale, the authors turn to developing lightweight yet more powerful models for easier deployment. \textbf{Mengzi-T5} was trained on Chinese Wikipedia, Chinese News, and
Common Crawl and the total size of the pre-training corpus is $300$G.

$\textbf{mBART}$~\citep{liu-etal-2020-multilingual-denoising} is a multi-lingual variant of BART, which is a sequence-to-sequence denoising auto-encoder. $\textbf{mBART}$ is pre-trained on large-scale monolingual corpora with the BART~\citep{lewis-etal-2020-bart} pre-training objective. The model performs extremely well in machine translation tasks and can be generalized to languages that are not in the pre-training corpora.

\textbf{C-BART}~\citep{shao2021cpt} is the Chinese version of BART. Compared with $\textbf{mBART}$, the model was pre-trained only on a Chinese corpus. The model shows superior performance on keyword recognition tasks evaluated by the Rouge-L metric.

$\textbf{CPM}$\footnote{\url{https://github.com/OpenBMB/CPM-Live}} is the generative pre-trained model series provided by OpenBMB\footnote{\url{https://live.openbmb.org/en/}}. We choose three PLMs $\textbf{CPM}_\textbf{2.6B}$ (CPM-1~\citep{zhang2021cpm}), $\textbf{CPM}_\textbf{7B}$ (CPM-Live), and $\textbf{CPM}_\textbf{10B}$ (CPM-Ant) with increasing model sizes. The three models are trained with increasingly larger sizes of data and training computations.

\paragraph{Training Details.}
For each model, we follow the configuration recommended by the original papers. During training, we select the model checkpoint with the best performance on the development set and evaluate it on the test set. The maximum sequence length is $2048$ for $\textbf{mT0}_\textbf{BASE}$, $\textbf{mT5}_\textbf{BASE}$, and $\textbf{Mengzi-T5}_\textbf{BASE}$, $1024$ for $\textbf{mBART}_\textbf{LARGE}$, $512$ for $\textbf{C-BART}_\textbf{LARGE}$, and $3072$ for $\textbf{CPM}$. We truncate the input sequence if it exceeds the maximum sequence length of a PLM.

\section{Design Differences between WebGPT and \ourmodel}
\label{sec:difference}
\paragraph{Interface.}
Our interface supports slightly different actions than WebGPT. To begin with, we remove $1$ actions defined by WebGPT: \texttt{Find in Page: <text>}, which supports finding the next occurrence of \texttt{<text>} and scrolling to it. In our pilot studies, even if we give the annotators the options for this action, our annotators seldom execute them. Considering that it may be hard for our model to learn those extremely low-frequency actions, we do not include both actions in the final list of our actions.

Secondly, we modify the functionalities of the \texttt{Scroll} actions in WebGPT. Specifically, WebGPT merged any consecutive \texttt{Scroll Down} and \texttt{Scroll Up} actions made by humans into new actions \texttt{Scroll Down <?>} and \texttt{Scroll Up <?>}, where $?$ is the number of consecutive actions. These new actions are utilized by their models instead of the original \texttt{Scroll Down} and \texttt{Scroll Up} actions. Therefore, there exists a gap between what humans actually perform and what the model is allowed to execute. We contend that this gap could result in problems for behavior cloning. Specifically, humans perform consecutive \texttt{Scroll Down} actions because after each action, they carefully check the current window and find nothing useful. However, when merging consecutive actions, the intermediate observations would not be shown to the model, which makes decision making even more difficult.

Finally, we also implement a new \texttt{Merge} action to support merging two supporting facts into one. As mentioned before, \texttt{Merge} is specially designed for extracting a supporting fact that crosses the boundary of two windows. This action is critical to avoid recording fragmentary supporting facts. As shown in Figure~\ref{fig:action_pie}, \texttt{Merge} takes up a relatively large ($5.4\%$) percentage among all the actions, which is frequently executed by our annotators. This action makes it possible for our annotators to extract extremely long sequences as supporting facts.

\paragraph{Framework.}
WebGPT does not disclose the implementation details for both interactive web search and information synthesis (i.e., BC model in the original paper). In view of this, we propose our own framework from scratch, with several design choices not mentioned by WebGPT:

We decompose the web search process into $3$ distinct sub-tasks, i.e., action prediction, search query generation, and supporting fact extraction. We train $3$ modules for each sub-task, respectively. This decomposition allows us to evaluate three modules in isolation and gain a deeper understanding of the strengths and weaknesses of each module. Furthermore, it allows for flexibility in the system, as different modules can be updated or replaced independently.

For our synthesis model, instead of directly fine-tuning on the (\textit{question}, \textit{supporting fact}, \textit{answer}) data, we explore (1) how to teach the model to ignore irrelevant facts (\cref{sec:in_depth_analysis}). We achieve this goal by introducing noisy facts into the training data to explicitly force the model to ignore noisy facts, and (2) how to generate novel contents beyond the collected facts (\cref{sec:novel_generation}). We corrupt the training data by deleting partial supporting facts and forcing the model to generate novel content based on its pre-trained knowledge.

\paragraph{Evaluation.}
WebGPT only evaluates the \textbf{the whole} pipeline through human evaluation. In addition to the holistic pipeline evaluation (\cref{sec:exp_pipeline}), we also evaluate each \textbf{individual} module of our pipeline (\cref{sec:exp_sub_task}). To the best of our knowledge, this is the first work to decompose interactive web search into action prediction, search query generation, and supporting fact extraction, and design the evaluation metrics for the three sub-tasks. It should be noted that holistic evaluation requires manual inspection, which is time-consuming despite being more accurate. Additionally, the holistic evaluation can only be conducted through interaction with the interface, whereas the individual sub-task evaluation can be conducted locally (by feeding the ground truth $\mathcal{S}_t$ of the test data to each module). As such, individual sub-task evaluation is more flexible to implement, making it easier for hyper-parameter tuning, thus accelerating the development and iteration of the QA system. Besides, individual evaluation is more fine-grained, which helps us better understand the contribution of each part of the pipeline.

\paragraph{Analysis.}
In addition to evaluating the LFQA performance of our pipeline, we also conduct an in-depth analysis to understand the contribution of core design elements of our framework. In \cref{sec:in_depth_analysis}, we conduct ablation studies for the search model and the synthesis model, and a case study for the query module. We also show that our model indeed acquires humanoid behaviors when interacting with the search engine.

\section{Additional Experiments and Analyses}

\subsection{Generating Novel Contents v.s. Copying Supporting Facts}
\label{sec:novel_generation}

Another fascinating research question of our synthesis model is whether it could generate novel content based on its pre-trained knowledge. This ability is important especially when the collected facts are insufficient or fragmentary. Considering that copying the supporting facts and generating novel contents are often contradictory to each other, here we propose a method to flexibly strike a balance between both.

\paragraph{Framework.}
Specifically, we propose another way to corrupt the training data of the synthesis model. We split each collected fact into multiple sub-sentences according to punctuation and randomly erase part of these sub-sentences. We set a hyper-parameter $p \in [0,1]$, which denotes the probability of erasing a sub-sentence. A higher $p$ means more sub-sentences would be removed. After that, we concatenate the remaining sub-sentences into a new fact keeping the original order. Finally, we optimize the model to generate the human-annotated answer conditioned on the corrupted facts, i.e., maximizing:
\begin{equation}
\small
\begin{aligned}
    \mathcal{P}(Answer|\mathcal{Q}_0, \textsc{Corrupt}\{f_1, ..., f_{\text{N}}\}).
    \nonumber
\end{aligned}
\end{equation}
Since the corrupted facts are fragmentary, the model learns to reconstruct those missing sub-sentences relying on its pre-trained knowledge.

\begin{table}[!t]
  \centering
  \small
    \begin{tabular}{ccccccc}
    \toprule
    $p$ & $0.1$     & $0.2$     & $0.3$     & $0.4$     & $0.5$     & $1.0$ \\
    \midrule
    \textsc{Novelty} & $0.06$ & $0.12$ & $0.16$ & $0.29$ & $0.41$ & $0.83$ \\
    Length & $256$ & $216$   & $206$   & $201$ & $193$ & $126$ \\
    \bottomrule
    \end{tabular}
  \caption{Results when the training data of synthesis model is corrupted with different $p$. We report two metrics for the generated sequence: \textsc{Novelty} and Length.}
  \label{tab:novel_exp}
\end{table}

\paragraph{Settings.}
We experiment with $\textbf{CPM}_\textbf{7B}$ and follow most of the settings in \cref{sec:exp_sub_task}. We test when different $p$ is applied to corrupt the training data. Ideally, a higher $p$ encourages the model to generate more novel content instead of copying the supporting facts. Specifically, we choose $p$ from $\{0.1, 0.2, 0.3, 0.4, 1.0\}$, where $1.0$ means the model sees no supporting facts but is required to generate all the tokens in the annotated answer.

During the evaluation, we feed \textbf{the original intact supporting facts} to the trained synthesis model. For evaluation metrics, we follow \citet{welleck2019neural} to test the percentage of n-grams in the generated sequence that do not exist in the supporting facts, i.e., 
\begin{equation}
\small
\begin{aligned}
    \textsc{Novelty}_\text{n} = \frac{|\text{unique generated n-grams}|}{|\text{total n-grams in supporting facts}|}.
    \nonumber
\end{aligned}
\end{equation}
The final novelty metric is defined as the average of $\textsc{Novelty}_\text{2}$, $\textsc{Novelty}_\text{3}$, and $\textsc{Novelty}_\text{4}$, i.e.,
\begin{equation}
\small
\begin{aligned}
    \textsc{Novelty} = \frac{1}{3} (\textsc{Novelty}_\text{2} + \textsc{Novelty}_\text{3} + \textsc{Novelty}_\text{4}).
    \nonumber
\end{aligned}
\end{equation}
Besides \textsc{Novelty}, we also record the number of generated tokens.

\paragraph{Results.} We derive from the results listed in Table~\ref{tab:novel_exp} that: (1) with $p$ increasing, the metric \textsc{Novelty} constantly becomes larger. This demonstrates that by deleting more content of the supporting facts during training, we gradually encourage the synthesis model to generate novel content based on its pre-trained knowledge, instead of copying the supporting facts. However, it should also be noted that the generated information that is not included in the collected facts may suffer from poor factual accuracy. We expect future work to mitigate this issue; (2) in addition, with $p$ increasing, the generated sequence tends to be shorter. This shows that only relying on the synthesis model cannot produce diverse, abundant, and informative contents, which emphasizes the importance of information retrieval in LFQA.

\section{Future Explorations}
\label{sec:future_work}
We expect future works to explore the following directions:

\paragraph{Efficient and Scalable Use.}
Despite the fascinating feature of interactive web search, such a process is inherently slower to execute than the conventional non-interactive retrieval process of open-domain QA. In this regard, we encourage further explorations in reducing the latency of our pipeline. Possible solutions include improving the speed and memory usage of the PLM.

\paragraph{Extension to Other Languages and Domains.} It would be interesting to extend the current approach to other languages beyond Chinese. Considering that the search engine supports multiple languages, our interface can be easily adapted to building benchmarks for other languages.

\paragraph{Leveraging the Reference Information.}
In addition to the annotated answers, we also require the annotators to record which supporting facts are referenced and leveraged in their answers. However, in this paper, we do not utilize this information when training our synthesis model. Intuitively, such information could guide the synthesis model to better organize existing supporting facts in a more coherent way, and to improve its ability in selecting important information and ignoring irrelevant noises.

\paragraph{Diversify the Interactive Elements.}
In this paper, we focus on supporting the mainstream web search actions for our users. It would interesting to explore incorporating more interactive elements into the interface, such as allowing the users to provide feedback on the retrieved information and supporting multimedia information retrieval. However, more actions also increase the difficulty of behavior cloning to a certain degree.

\paragraph{Improving Model Behavior from Human Feedbacks.}
WebGPT has demonstrated it is promising to use reinforcement learning from human feedback (RLHF)~\citep{stiennon2020learning} to improve the quality of the generated answers. RLHF can also be used for improving the search model's web search behavior, and make it collect more diverse and relevant supporting facts. Our provided environment can be utilized by researchers to study RLHF in the future.

\end{document}